\documentclass{article}

\usepackage{iclr2017_conference,times}

\usepackage[utf8]{inputenc} 
\usepackage[T1]{fontenc}    
\usepackage{hyperref}       
\usepackage{url}            
\usepackage{booktabs}       
\usepackage{amsfonts}       
\usepackage{nicefrac}       
\usepackage{microtype}      

\usepackage{algorithm2e}
\usepackage{amsmath}
\usepackage{amsthm}
\usepackage{amsfonts}
\usepackage{amssymb}
\usepackage{mathpartir}
\usepackage{mathtools}
\usepackage{mathrsfs}
\usepackage{mathtools}
\usepackage{siunitx}
\usepackage{url}

\title{Low-rank passthrough neural networks}

\author{Antonio Valerio Miceli Barone  \thanks{Work partially done while affiliated with University of Pisa.} \\
School of Informatics \\
The University of Edinburgh \\
\texttt{amiceli@inf.ed.ac.uk} \\
}

\begin{document}

\maketitle

\begin{abstract}
Deep learning consists in training neural networks to perform computations that sequentially unfold in many steps over a time dimension or an intrinsic depth dimension. For large depths, this is usually accomplished by specialized network architectures that are designed to mitigate the vanishing gradient problem, e.g. LSTMs, GRUs, Highway Networks and Deep Residual Networks, which are based on a single structural principle: the state passthrough.  
We observe that these "Passthrough Networks" architectures enable the decoupling of the network state size from the number of parameters of the network, a possibility that is exploited in some recent works but not thoroughly explored.  
In this work we propose simple, yet effective, low-rank and low-rank plus diagonal matrix parametrizations for Passthrough Networks which exploit this decoupling property, reducing the data complexity and memory requirements of the network while preserving its memory capacity. We present competitive experimental results on several tasks, including a near state of the art result on sequential randomly-permuted MNIST classification, a hard task on natural data.
\end{abstract}

\section{Overview}

Deep neural networks can perform non-trivial computations by the repeated the application of parametric non-linear transformation layers to vectorial (or, more generally, tensorial) data. This staging of many computation steps can be done over a time dimension for tasks involving sequential inputs or outputs of varying length, yielding a \textit{recurrent neural network}, or over an intrinsic circuit depth dimension, yielding a \textit{deep feed-forward neural network}, or both. Training these deep models is complicated by the \textit{exploding} and \textit{vanishing} gradient problems \citep{Hochreiter1991, Bengio1994}.

Various network architectures have been proposed to ameliorate the vanishing gradient problem in the recurrent setting, such as the LSTM \citep{Hochreiter1997,Graves2005}, the GRU \citep{Cho2014}, etc. These architectures led to a number of breakthroughs in different tasks in NLP, computer vision, etc. \citep{Graves2013, Cho2014a, Bahdanau2014, Vinyals2014, Iyyer2014}. Similar methods have also been applied in the feed-forward setting with architectures such as Highway Networks \citep{Srivastava2015}, Deep Residual Networks \citep{He2015}, and so on. All these architectures are based on a single structural principle which, in this work, we will refer to as the \textit{state passthrough}. We will thus refer to these architectures as \textit{Passthrough Networks}.

Another difficulty in training neural networks is the trade-off between the network representation power and its number of trainable parameters, which affects its data complexity during training in addition to its implementation memory requirements. On one hand, the number of parameters can be thought as the number of tunable "knobs" that need to be set to represent a function, on the other hand, it also constrains the size of the partial results that are propagated inside the network. In typical fully connected networks, a layer acting on a $n$-dimensional state vector has $O(n^2)$ parameters stored in one or more matrices, but there can be many functions of practical interest that are simple enough to be represented by a relatively small number of bits while still requiring some sizable amount of memory to be computed. Therefore, representing these functions on a fully connected neural network can be wasteful in terms of number of parameters. The full parameterization implies that, at each step, all the information in each state component can affect all the information in any state component at the next step. Classical physical systems, however, consist of spatially separated parts with primarily local interactions, long-distance interactions are possible but they tend to be limited by propagation delays, bandwidth and noise. Therefore it may be beneficial to bias our model class towards models that tend to adhere to these physical constraints by using a parametrization which reduces the number of parameters required to represent them. This can be accomplished by imposing some constraints on the $n \times n$ matrices that parametrize the state transitions. One way of doing this is to impose a convolutional structure on these matrices \citep{LeCun2004, Krizhevsky2012}, which corresponds to strict locality and periodicity constraints as in a cellular automaton. These constraints work well in certain domains such as vision, but may be overly restrictive in other domains.

In this work we observe that the state passthrough allows for a systematic decoupling of the network state size from the number of parameters: since by default the state vector passes mostly unaltered through the layers, each layer can be made simple enough to be described only by a small number of parameters without affecting the overall memory capacity of the network, effectively spreading the computation over the depth or time dimension of the network, but without making the network "thin". This has been exploited by some convolutional passthrough architectures \citep{Srivastava2015, He2015, Kaiser2015}, or architectures with addressable read-write memory \citep{Graves2014, Danihelka2016}.

In this work we propose simple but effective low-dimensional parametrizations that exploit this decoupling based on low-rank or low-rank plus diagonal matrix decompositions. Our approach extends the LSTM architecture with a single projection layer by \citet{Sak2014} which has been applied to speech recognition, natural language modeling \citep{Jozefowicz2016}, video analysis \citep{Sun2015}, etc. We provide experimental evaluation of our approach on GRU and LSTM architectures on various machine learning tasks, including a near state of the art result for the hard task of sequential randomly-permuted MNIST image recognition \citep{Le2015}.

\section{Model}
\label{SEC:MODEL}
A neural network can be described as a dynamical system that transforms an input $u$ into an output $y$ over multiple time steps $T$. At each step $t$ the network has a $n$-dimensional state vector $x(t) \in \mathcal{R}^n$ defined as
\begin{equation}
x(t) =
\begin{cases}
 in(u, \theta) & \text{ if } t= 0\\ 
 f(x(t-1), t, u, \theta) & \text{ if } t \geq 1
\end{cases}
\end{equation}
where $in$ is a \textit{state initialization function}, $f$ is a \textit{state transition function} and $\theta \in \mathcal{R}^k$ is vector of trainable parameters. The output $y = out(x(0:T), \theta)$ is generated by an \textit{output function} $out$, where $x(0:T)$ denotes the whole sequence of states visited during the execution.  
In a feed-forward neural network with constant hidden layer width $n$, the input $u \in \mathcal{R}^m$ and the output $y \in \mathcal{R}^l$ are vectors of fixed dimension $m$ and $l$ respectively, $T$ is a model hyperparameter.  
In a recurrent neural network the input $u$ is typically a list of $T$ $m$-dimensional vectors $u(t) \in \mathcal{R}^m$ for $t \in 1, \dots, T$ where $T$ is variable, the output $y$ is either a single $l$-dimensional vector or a list of $T$ such vectors. Other neural architectures, such as "seq2seq" transducers without attention \citep{Cho2014a}, can be also described within this framework.

\subsection{Passthrough networks}
\label{SEC:MODEL:PN}

Passthrough networks can be defined as networks where the state transition function $f$ has a special form such that, at each step $t$ the state vector $x(t)$ (or a sub-vector $\hat{x}(t)$) is propagated to the next step modified only by some (nearly) linear, element-wise transformation.

Let the state vector $x(t) \equiv (\hat{x}(t), \tilde{x}(t))$ be the concatenation of $\hat{x}(t) \in \mathcal{R}^{\hat{n}}$ and $\tilde{x}(t) \in \mathcal{R}^{\tilde{n}}$ with $\hat{n} + \tilde{n} = n$ (where $\tilde{n}$ can be equal to zero). We define a network to have a \textit{state passthrough} on $\hat{x}$ if $\hat{x}$ evolves as
\begin{equation}
\hat{x}(t) = f_{\pi}(x(t-1), t, u, \theta) \odot f_{\tau}(x(t-1), t, u, \theta) + \hat{x}(t-1) \odot f_{\gamma}(x(t-1), t, u, \theta)
\end{equation}
where $f_{\pi}$ is the \textit{next state proposal function}, $f_{\tau}$ is the \textit{transform function}, $f_{\gamma}$ is the \textit{carry function} and $\odot$ denotes  element-wise vector multiplication. The rest of the state vector $\tilde{x}(t)$, if present, evolves according to some other function $\tilde{f}$. In practice $\tilde{x}(t)$ is only used in LSTM variants, while in other passthrough architectures $\hat{x}(t) = x(t)$.

\begin{figure}
\setlength\tabcolsep{10.0pt}
\begin{tabular}{l|l}
\includegraphics[scale=0.5]{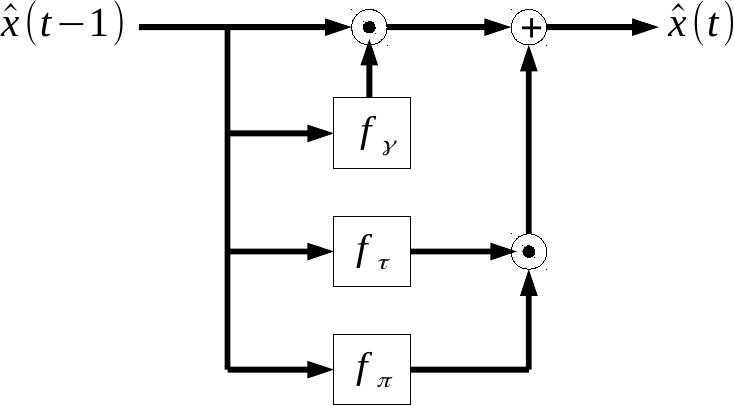} 
&
\includegraphics[scale=0.35]{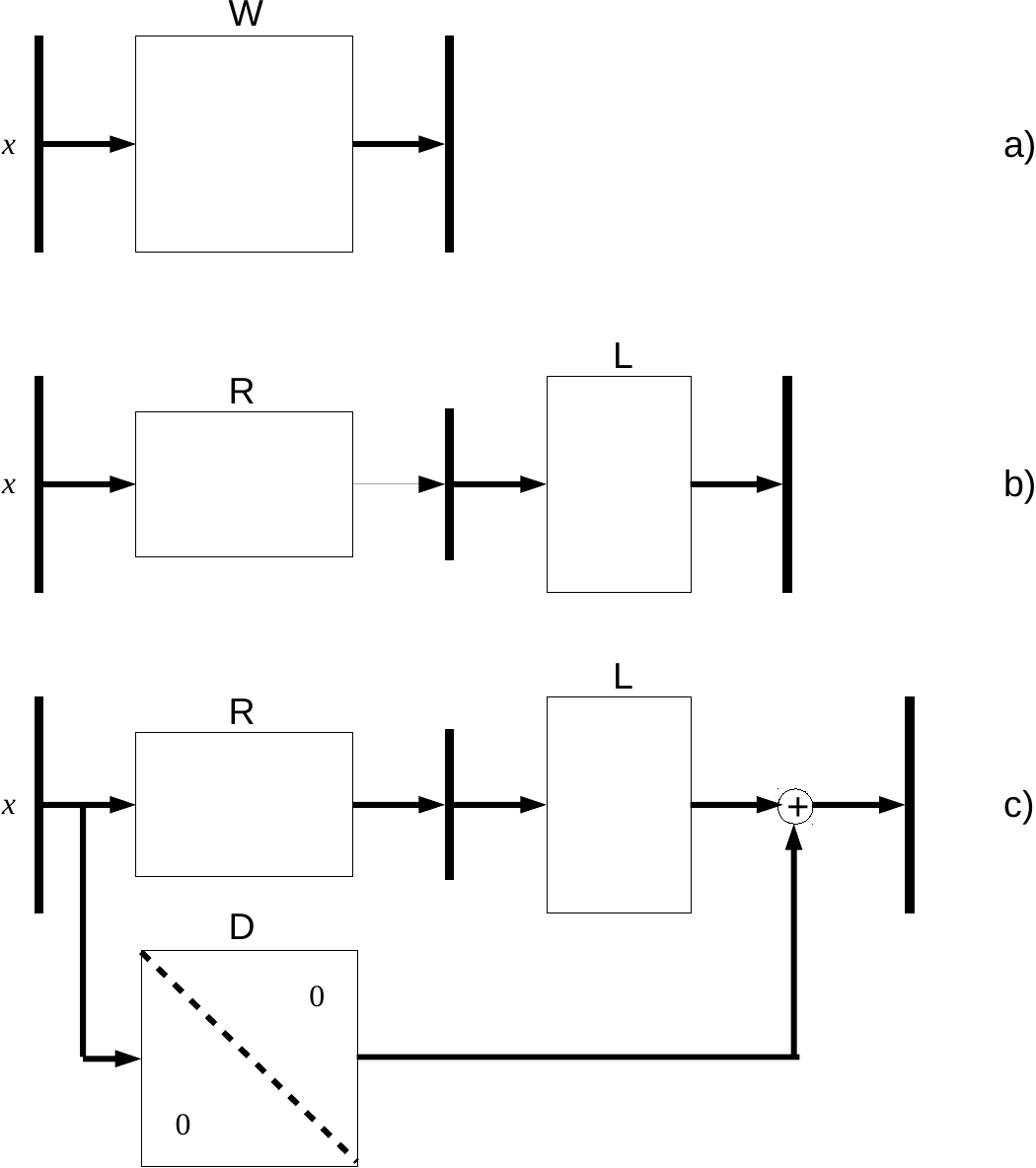} 
\end{tabular}
\caption{Left: Generic state passthrough hidden layer, optional non-passthrough state $\tilde{x}(t)$ and per-timestep input $u(t)$ are not shown. Right: a) Full matrix parametrization. b) Low-rank parametrization. c) Low-rank plus diagonal parametrization.}
\label{FIG:PASSTRHOUGH_LRDPARAM}
\end{figure}

As concrete example, we can describe a fully connected Highway Network as
\begin{equation}
\begin{aligned}
f_{\pi}(x(t-1), t, u, \theta)    &= g(\theta^{(W_{\pi})}_t \cdot x(t-1) + \theta^{(b_{\pi})}_t) \\
f_{\tau}(x(t-1), t, u, \theta)   &= \sigma(\theta^{(W_{\tau})}_t \cdot x(t-1) + \theta^{(b_{\tau})}_t) \\
f_{\gamma}(x(t-1), t, u, \theta) &=  1^{\otimes n} - f_{\tau}(x(t-1), t, u, \theta)
\end{aligned}
\label{EQ:MODEL:PN:HIGHWAY}
\end{equation}
where $g$ is an element-wise activation function, usually the ReLU \citep{Glorot2011} or the hyperbolic tangent, $\sigma$ is the element-wise logistic sigmoid, and $\forall t \in 1, \dots, T$, the parameters $\theta^{(W_{\pi})}_t$ and $\theta^{(W_{\tau})}_t$ are matrices in $\mathcal{R}^{n \times n}$ and $\theta^{(b_{\pi})}_t$ and $\theta^{(b_{\pi})}_t$ are vectors in $\mathcal{R}^n$. Dependence on the input $u$ occurs only through the initialization function, which is model-specific and is omitted here, as is the output function.

\subsection{Low-rank passthrough networks}
\label{SEC:MODEL:LRPN}

In fully connected architectures there are $n \times n$ matrices that act on the state vector, such as the $\theta^{(W_{\pi})}_t$ and $\theta^{(W_{\tau})}_t$ matrices of the Highway Network of eq. \ref{EQ:MODEL:PN:HIGHWAY}. Each of these matrices has $n^2$ entries, thus for large $n$, the entries of these matrices can make up the majority of independently trainable parameters of the model. As discussed in the previous section, this parametrization can be wasteful. We impose a low-rank constraint on these matrices. This is easily accomplished by rewriting each of these matrices as the product of two matrices where the inner dimension $d$ is a model hyperparameter.

For instance, in the case of the Highway Network of eq. \ref{EQ:MODEL:PN:HIGHWAY} we can redefine $\forall t \in 1, \dots, T$
\begin{equation}
\begin{aligned}
\theta^{(W_{\pi})}_t  &= \theta^{(L_{\pi})}_t \cdot \theta^{(R_{\pi})}_t \\
\theta^{(W_{\tau})}_t &= \theta^{(L_{\tau})}_t \cdot \theta^{(R_{\tau})}_t
\end{aligned}
\label{EQ:MODEL:PN:LRHIGHWAY}
\end{equation}
where $\theta^{(L_{\pi})}_t, \theta^{(L_{\tau})}_t \in \mathcal{R}^{n \times d}$ and $\theta^{(R_{\pi})}_t, \theta^{(R_{\tau})}_t \in \mathcal{R}^{d \times n}$. When $d < n/2$ this result in a reduction of the number of trainable parameters of the model.

Even when $n/2 \leq d < n$, while the total number of parameter increases, the number of degrees of freedom of the model still decreases, because low-rank factorization are unique only up to arbitrary $d \times d$ invertible matrices, thus the number of independent degrees of freedom of a low-rank layer is $2nd - d^2$. However, we don't know whether the training optimizers can exploit this kind of redundancy, thus in this work we restrict to low-rank parametrizations where the number of parameters is strictly reduced.

This low-rank constraint can be thought as a bandwidth constraint on the computation performed at each step: the $R$ matrices first project the state into a smaller subspace, extracting the information needed for that specific step, then the $L$ matrices project it back to the original state space, spreading the selected information to all the state components that need to be updated. A similar approach has been proposed for the LSTM architecture by \citet{Sak2014}, although they force the $R$ matrices to be the same for all the functions of the state transition, while we allow each parameter matrix to be parametrized independently by a pair of $R$ and $L$ matrices.

Low-rank passthrough architectures are universal in that they retain the same representation classes of their parent architectures. This is trivially true if the inner dimension $d$ is allowed to be $O(n)$ in the worst case, and for some architectures even if $d$ is held constant. For instance, it is easily shown that for any Highway Network with state size $n$ and $T$ hidden layers and for any $\epsilon > 0$, there exist a Low-rank Highway Network with $d = 1$, state size at most $2n$ and at most $nT$ layers that computes the same function within an $\epsilon$ margin of error.

\subsection{Low-rank plus diagonal passthrough networks}
\label{SEC:MODEL:LRDPN}

As we show in the experimental section, on some tasks the low-rank constraint may prove to be excessively restrictive if the goal is to train a model with fewer parameters than one with arbitrary matrices. A simple extension is to add to each low-rank parameter matrix a diagonal parameter matrix, yielding a matrix that is full-rank but still parametrized in a low-dimensional space. For instance, for the Highway Network architecture we modify eq. \ref{EQ:MODEL:PN:LRHIGHWAY} to
\begin{equation}
\begin{aligned}
\theta^{(W_{\pi})}_t  &= \theta^{(L_{\pi})}_t \cdot \theta^{(R_{\pi})}_t + \theta^{(D_{\pi})}_t \\
\theta^{(W_{\tau})}_t &= \theta^{(L_{\tau})}_t \cdot \theta^{(R_{\tau})}_t + \theta^{(D_{\tau})}_t
\end{aligned}
\label{EQ:MODEL:PN:LRDHIGHWAY}
\end{equation}
where $\theta^{(D_{\pi})}_t, \theta^{(D_{\tau})}_t \in \mathcal{R}^{n \times n}$ are trainable diagonal parameter matrices.

It may seem that adding diagonal parameter matrices is redundant in passthrough networks. After all, the state passthrough itself can be considered as a diagonal matrix applied to the state vector, which is then additively combined to the new proposed state computed by the $f_{\pi}$ function. However, since the state passthrough completely skips over all non-linear activation functions, these formulations are not equivalent. In particular, the low-rank plus diagonal parametrization may help in recurrent neural networks which receive input at each time step, since they allow each component of the state vector to directly control how much input signal is inserted into it at each step. We demonstrate the effectiveness of this model in the sequence copy and sequential MNIST tasks described in the experiments section.

\section{Experiments}

The main content of this section reports several experiments on Low-rank and Low-rank plus diagonal GRUs, and an experiment using these parametrizations on a LSTM for language modeling.

A preliminary experiment on Low-rank Highway Networks on the MNIST dataset is reported in appendix \ref{APPENDIXLRHN}.

We applied the Low-rank and Low-rank plus diagonal GRU architectures to a subset of sequential benchmarks described in the Unitary Evolution Recurrent Neural Networks article by \citet{Arjovsky2015}, specifically the memory task, the addition task and the sequential randomly permuted MNIST task. For the memory tasks, we also considered two different variants proposed by \citet{Danihelka2016} and \citet{Henaff2016} which are hard for the uRNN architecture. We chose to compare against the uRNN architecture because it set state of the art results in terms of both data complexity and accuracy and because it is an architecture with similar design objectives as low-rank passthrough architectures, namely a low-dimensional parametrization and the mitigation of the vanishing gradient problem, but it is based on quite different principles.

The GRU architecture \citep{Cho2014} is a passthrough recurrent neural network defined as
\begin{equation}
\begin{aligned}
in(u, \theta)           &= \theta_{in} \\
f_{\omega}(x(t-1), t, u, \theta)    &= \sigma(\theta^{U_{\omega}} \cdot u(t) + \theta^{(W_{\omega})} \cdot x(t-1) + \theta^{(b_{\omega})}) \\
f_{\gamma}(x(t-1), t, u, \theta)      &= \sigma(\theta^{U_{\gamma}} \cdot u(t) + \theta^{(W_{\gamma})} \cdot x(t-1) + \theta^{(b_{\gamma})}) \\
f_{\tau}(x(t-1), t, u, \theta) &=  1^{\otimes n} - f_{\gamma}(x(t-1), t, u, \theta) \\
f_{\pi}(x(t-1), t, u, \theta)    &= \text{tanh}(\theta^{U_{\pi}} \cdot u(t) + \theta^{(W_{\pi})} \cdot (x(t-1) \odot f_{\omega}(x(t-1), t, u, \theta)) + \theta^{(b_{\pi})}) 
\end{aligned}
\label{EQ:EXPERIMENTS:GRU}
\end{equation}

We turn this architecture into the Low-rank GRU architecture by redefining each of the $\theta^W$ matrices as the product of two matrices with inner dimension $d$. For the memory tasks, which turned out to be difficult for the low-rank parametrization, we also consider the low-rank plus diagonal parametrization. We also applied the low-rank plus diagonal parametrization in the sequential permuted MNIST task and a character-level language modeling task on the Penn Treebank corpus. For the language modeling task, we also experimented with Low-rank plus diagonal LSTMs. Refer to appendix \ref{APPENDIXLRGRU} for model details.

\begin{figure}
\setlength\tabcolsep{0.0pt}
\begin{tabular}{ll}
\includegraphics[scale=0.38]{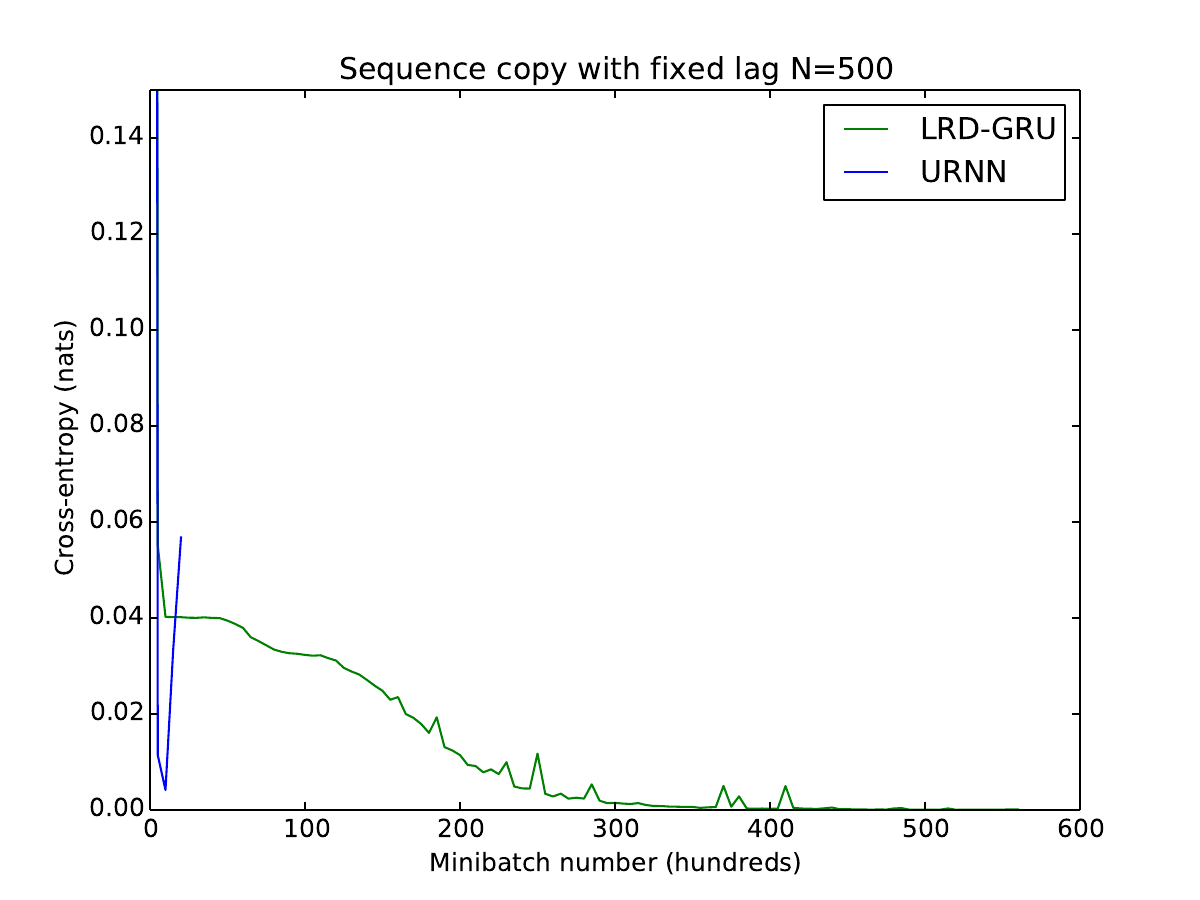}
&
\includegraphics[scale=0.38]{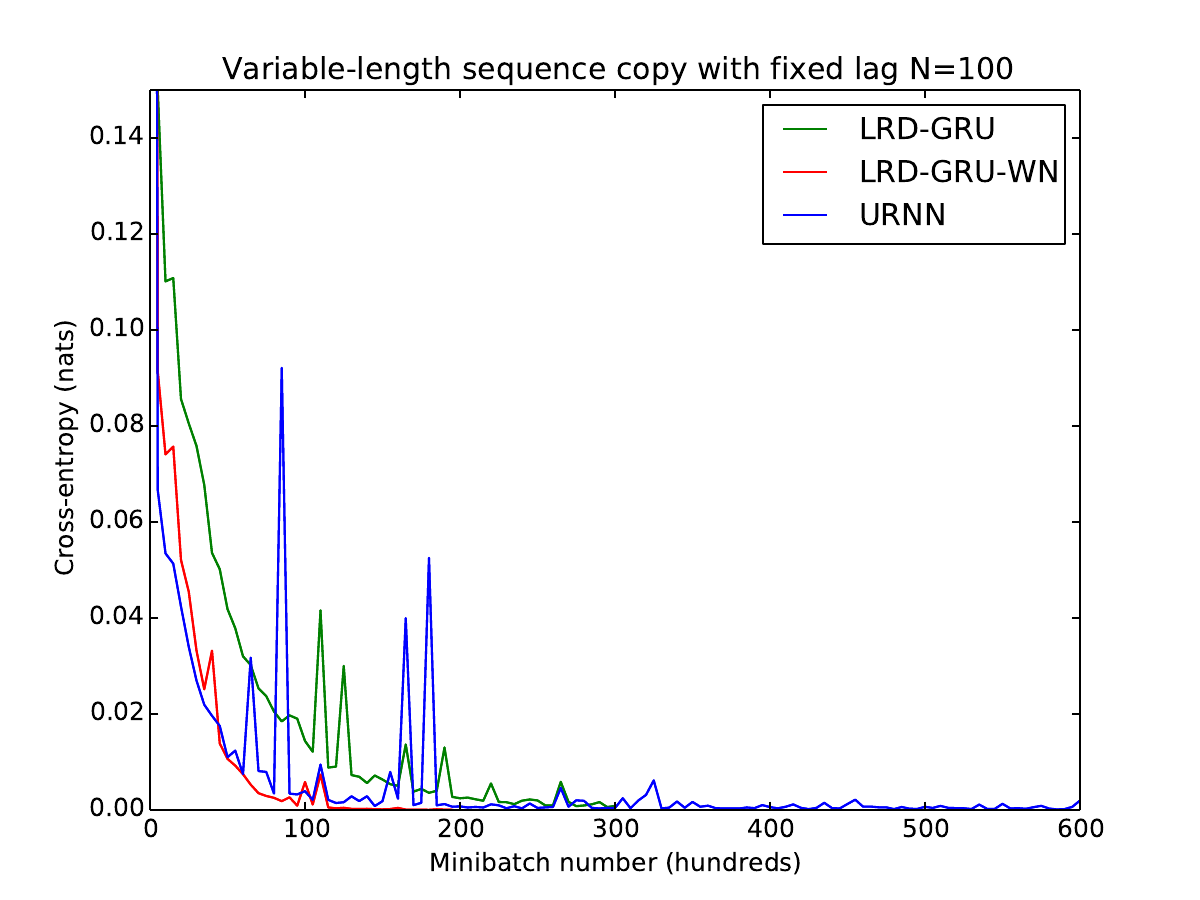} 
\\
\includegraphics[scale=0.38]{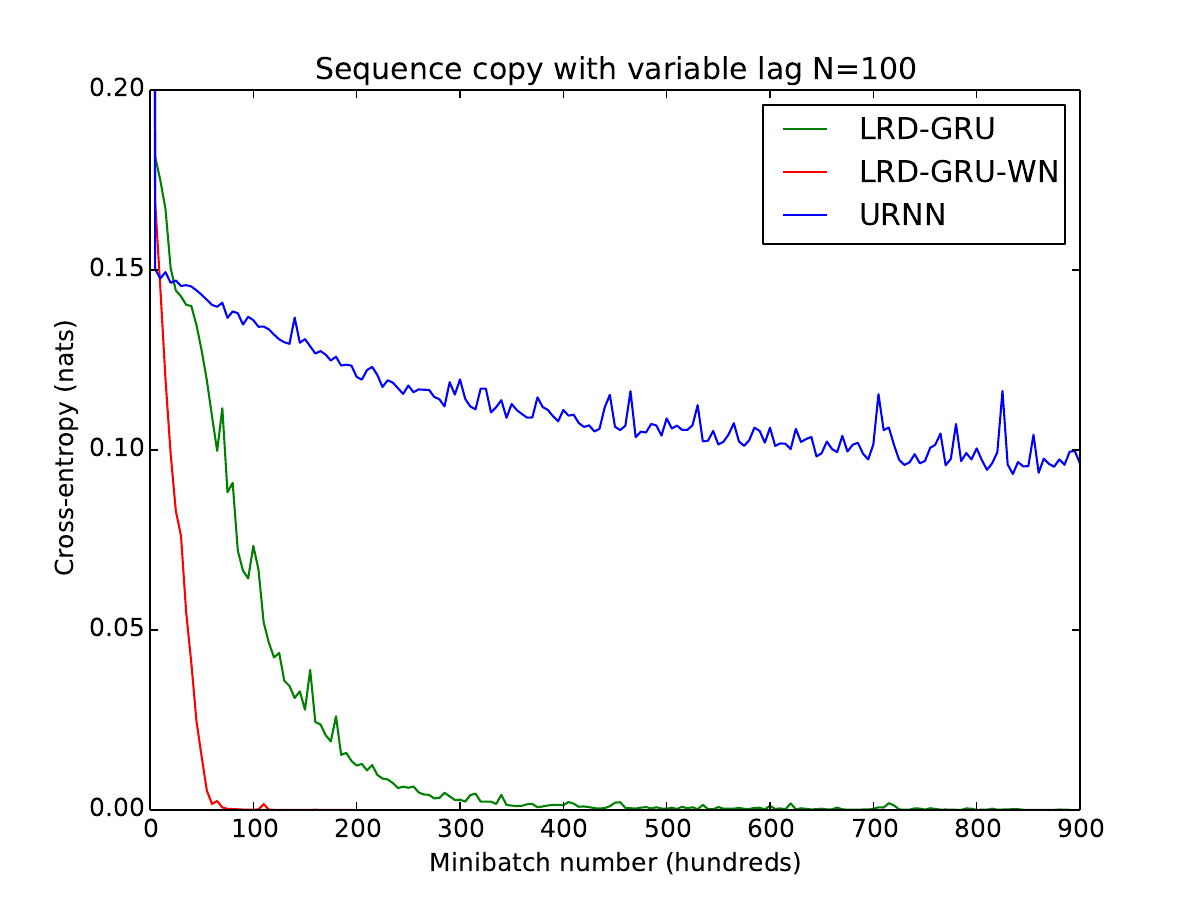} 
&
\includegraphics[scale=0.38]{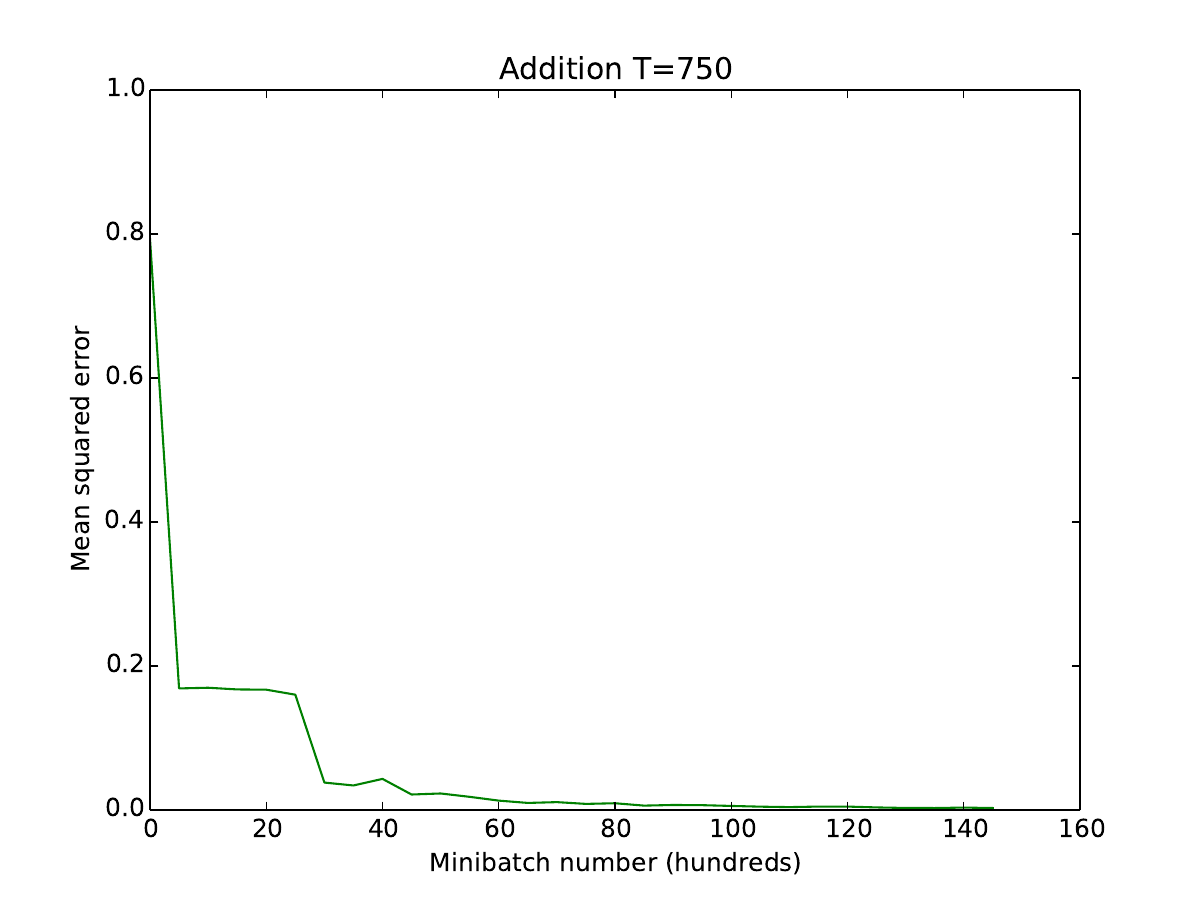} 
\\
\includegraphics[scale=0.38]{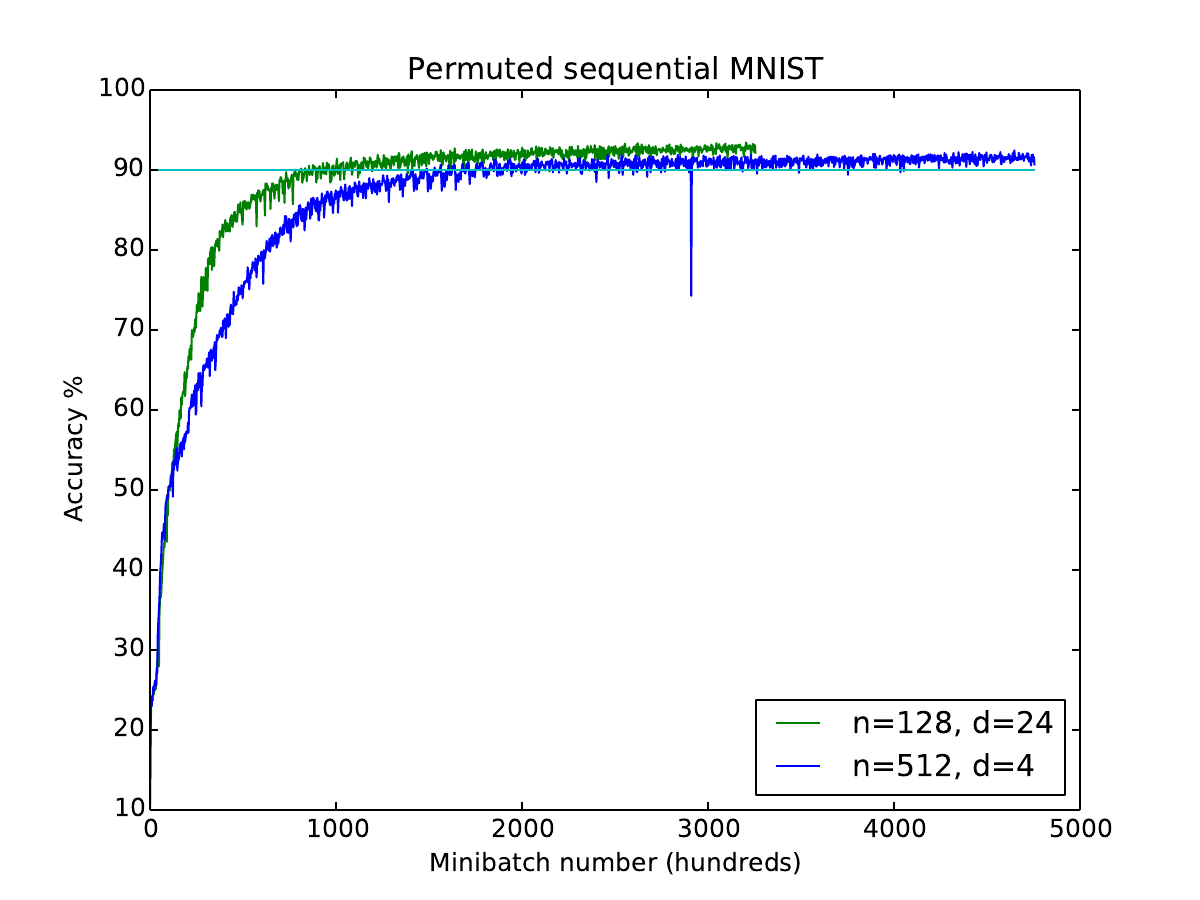} 
&
\includegraphics[scale=0.38]{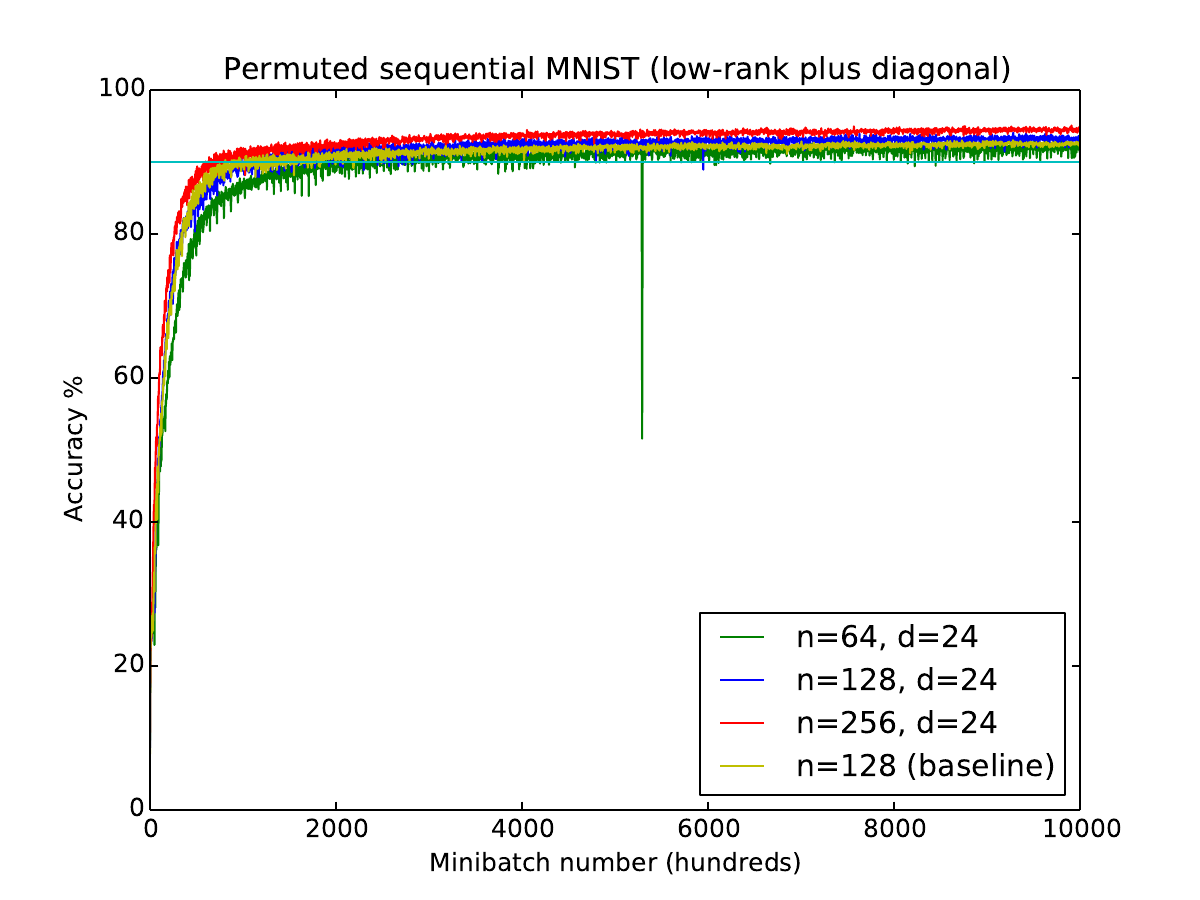} 
\end{tabular}
\caption{Top row and middle left: Low-rank plus diagonal GRU and uRNN on the sequence copy tasks, cross-entropy on validation set. Middle right: Low-rank GRU on the addition task, mean squared error on validation set. Bottom row: Low-rank GRU (left) and Low-rank plus diagonal GRU (right) on the permuted sequential MNIST task, accuracy on validation set, horizontal line indicates 90\% accuracy.}
\label{FIG:LRDGRU}
\end{figure}

\subsubsection{Memory task}
The input of an instance of this task is a sequence of $T=N+20$ discrete symbols in a ten symbol alphabet ${a_i : i \in 0, \dots 9}$, encoded as one-hot vectors. The first $10$ symbols in the sequence are "data" symbols i.i.d. sampled from $a_0, \dots, a_7$, followed by $N-1$ "blank" $a_8$ symbols, then a distinguished "run" symbol $a_9$, followed by $10$ more "blank" $a_8$ symbols. The desired output sequence consists of $N+10$ "blank" $a_8$ symbols followed by the $10$ "data" symbols as they appeared in the input sequence. Therefore the model has to remember the $10$ "data" symbol string over the temporal gap of size $N$, which is challenging for a recurrent neural network when $N$ is large. In our experiment we set $N=500$, which is the hardest setting explored in the uRNN work. The training set consists of $100,000$ training examples and $10,000$ validation/test examples. The architecture is described by eq. \eqref{EQ:EXPERIMENTS:GRU}, with an additional output layer with a dense $n \times 10$ matrix followed a (biased) softmax. We train to minimize the cross-entropy loss. 

We were able to solve this task using a GRU with full recurrent matrices with state size $n=128$, learning rate $\num{1e-3}$, mini-batch size $20$, initial bias of the carry functions (the "update" gates) $4.0$, however this model has many more parameters, nearly $50,000$ in the recurrent layer only, than the uRNN work which has about $6,500$, and it converges much more slowly than the uRNN. We were not able to achieve convergence with a pure low-rank model without exceeding the number of parameters of the fully connected model, but we achieved fast convergence with a low-rank plus diagonal model with $d=50$, with other hyperparameters set as above. This model has still more parameters ($39,168$ in the recurrent layer, $41,738$ total) than the uRNN model and converges more slowly but still reasonably fast, reaching test cross-entropy $< \num{1e-3}$ nats and almost perfect classification accuracy in less than $35,000$ updates.

In order to obtain a fair comparison, we also train a uRNN model with state size $n=721$, resulting in approximately the same number of parameters as the low-rank plus diagonal GRU models. This model very quickly reaches perfect accuracy on the training set in less than $2,000$ updates, but overfits w.r.t. the test set.

We also consider two variants of this task which are difficult for the uRNN model. For both these tasks we used the same settings as above except that the task size parameter is set at $N=100$ for consistency with the works that introduced these variants. In the variant of \citet{Danihelka2016}, the length of the sequence to be remembered is randomly sampled between $1$ and $10$ for each sequence. They manage to achieve fast convergence with their Associative LSTM architecture with $65,505$ parameters, and slower convergence with standard LSTM models. Our low-rank plus diagonal GRU architecture, which has less parameters than their Associative LSTM, performs comparably or better, reaching test cross-entropy $< \num{1e-3}$ nats and almost perfect classification accuracy in less than $30,000$ updates. In the variant of \citet{Henaff2016}, the length of the sequence to be remembered is fixed at $10$ but the model is expected to copy it after a variable number of time steps randomly chosen, for each sequence, between $1$ and $N=100$. The authors achieve slow convergence with a standard LSTM model, while our low-rank plus diagonal GRU architecture achieves fast convergence, reaching test cross-entropy $< \num{1e-3}$ nats and almost perfect classification accuracy in less than $38,000$ updates, and perfect test accuracy in $87,000$ updates.

We further train uRNN models with state size $n=721$ on these variants of the memory task. We found that the uRNN learns faster than the low-rank plus diagonal GRU on the variable length, fixed lag task \citep{Danihelka2016} but fails to converge within our training time limit on the fixed length, variable lag task \citep{Henaff2016}.

Training the low-rank plus diagonal GRU on these tasks incurs sometimes in numerical stability problems as discussed in appendix \ref{APPENDIXLRGRU}. In order to systemically address these issues, we also trained models with weight normalization \citep{Salimans2016} and weight row max-norm constraints. These models turned out to be more stable and in fact converge faster, performing on par with the uRNN on the variable length, fixed lag task. 

Training curves are shown in figure \ref{FIG:LRDGRU} (top and middle left).

\subsubsection{Addition task}
For each instance of this task, the input sequence has length $T$ and consists of two real-valued components, at each step the first component is independently sampled from the interval $[0, 1]$ with uniform probability, the second component is equal to zero everywhere except at two randomly chosen time step, one in each half of the sequence, where it is equal to one. The result is a single real value computed from the final state which we want to be equal to the sum of the two elements of the first component of the sequence at the positions where the second component was set at one. In our experiment we set $T=750$.

The training set consists of $100,000$ training examples and $10,000$ validation/test examples. We use a Low-rank GRU with $2 \times n$ input matrix, $n \times 1$ output matrix and (biased) identity output activation. We train to minimize the mean squared error loss. We use state size $n=128$, maximum rank $d=24$. This results in approximately $6,140$ parameters in the recurrent hidden layer. Learning rate was set at $\num{1e-3}$, mini-batch size $20$, initial bias of the carry functions (the "update" gates) was set to $4$.

We trained on $14,500$ mini-batches, obtaining a mean squared error on the test set of $0.003$, which is a better result than the one reported in the uRNN article, in terms of training time and final accuracy. The training curve is shown in figure \ref{FIG:LRDGRU} (middle right).

\subsubsection{Sequential MNIST task}

This task consists of handwritten digit classification on the MNIST dataset with the caveat that the input is presented to the model one pixel value at time, over $T=784$ time steps. To further increase the difficulty of the task, the inputs are reordered according to a random permutation (fixed for all the task instances).

We use Low-rank and Low-rank plus diagonal GRUs with $1 \times n$ input matrix, $n \times 10$ output matrix and (biased) softmax output activation. Learning rate was set at $\num{5e-4}$, mini-batch size $20$, initial bias of the carry functions (the "update" gates) was set to $5$. 

\begin{table}[t]
\caption{Sequential permuted MNIST results}
\centering
\begin{tabular}{lrrrrr}
\toprule
Architecture & state size & max rank & params & val. accuracy & test accuracy \\
\midrule
Baseline GRU & 128 & - & $51.0$ k & $93.0\%$ & $92.8\%$ \\
Low-rank GRU & 128 & 24 & $20.2$ k & $93.4\%$ & $91.8\%$ \\
Low-rank GRU & 512 & 4 & $19.5$ k & $92.5\%$ & $91.3\%$ \\
Low-rank plus diag. GRU & 64 & 24 & $10.3$ k & $93.1\%$ & $91.9\%$ \\
Low-rank plus diag. GRU & 128 & 24 & $20.6$ k & $94.1\%$ & $93.5\%$ \\
Low-rank plus diag. GRU & 256 & 24 & $41.2$ k & $\mathbf{95.1\%}$ & $\mathbf{94.7\%}$ \\
\bottomrule
\end{tabular}
\label{TABLE:SEQMNIST}
\end{table}

Results are presented in table \ref{TABLE:SEQMNIST} and training curves are shown in figure \ref{FIG:LRDGRU} (bottom row). All these models except the one with the most extreme bottleneck ($n=512,d=4$) exceed the reported uRNN test accuracy of $91.4\%$, although they converge more slowly (hundred of thousands updates vs. tens of thousands of the uRNN). Also note that the low-rank plus diagonal GRU is more accurate than the full rank GRU with the same state size, while the low-rank GRU is slightly less accurate (in terms of test accuracy), indicating the utility of the diagonal component of the parametrization for this task.

These are on par with more complex architectures with time-skip connections \citep{zhang2016} (reported test set accuracy $94.0\%$). To our knowledge, at the time of this writing, the best result on this task is the LSTM with recurrent batch normalization by \citet{Cooijmans2016} (reported test set accuracy $95.2\%$). The architectural innovations of these works are orthogonal to our own and in principle they can be combined to it.

\subsubsection{Character-level language modeling task}

This standard benchmark task consist of predicting the probability of the next character in a sentence after having observed the previous charters. Similar to \citet{Zaremba2014}, we use the Penn Treebank English corpus, with standard training, validation and test splits.

As a baseline we use a single layer GRU either with no regularization or regularized with   Bayesian recurrent dropout \citep{Gal2015}. Refer to appendix \ref{APPENDIXLRGRU} for details.

\begin{table}[t]
\caption{Character-level language modeling results}
\centering
\begin{tabular}{lllrrrrr}
\toprule
Architecture & dropout & tied & state size & max rank & params & test per-char. perplexity \\
\midrule
Baseline GRU & No & - & 1000 & - & $3.11$ M & $2.96$ \\
Baseline GRU & Yes & - & 1000 & - & $3.11$ M & $2.92$ \\
Baseline GRU & Yes & - & 3298 & - & $33.0$ M & $2.77$ \\
Baseline LSTM & Yes & - & 1000 & - & $4.25$ M & $2.92$ \\
Low-rank plus diag. GRU  & No & No & 1000 & 64 & $0.49$ M & $2.92$\\
Low-rank plus diag. GRU  & No & No & 3298 & 128 & $2.89$ M & $2.95$\\
Low-rank plus diag. GRU  & Yes & No & 3298 & 128 & $2.89$ M & $2.86$\\
Low-rank plus diag. GRU  & Yes & No & 5459 & 64 & $2.69$ M & $2.82$\\
Low-rank plus diag. GRU  & Yes & Yes & 5459 & 64 & $1.99$ M & $2.81$\\
Low-rank plus diag. GRU  & No & Yes & 1000 & 64 & $0.46$ M & $2.90$\\
Low-rank plus diag. GRU  & Yes & Yes & 4480 & 128 & $2.78$ M & $2.86$\\
Low-rank plus diag. GRU  & Yes & Yes & 6985 & 64 & $2.54$ M & $\mathbf{2.76}$\\
Low-rank plus diag. LSTM & Yes & No & 1740 & 300 & $4.25$ M & $2.86$\\
\bottomrule
\end{tabular}
\label{TABLE:CHARLM}
\end{table}

In our experiments we consider the low-rank plus diagonal parametrization, both with tied and untied projection matrices. We set the state size and maximum rank to either reduce the total number of parameters compared to the baselines or to keep the number of parameters approximately the same while increasing the memory capacity. Results are shown in table \ref{TABLE:CHARLM}.

Our low-rank plus diagonal parametrization reduces the model per-character perplexity (the base-2 exponential of the bits-per-character entropy). Both the tied and untied versions perform equally when the state size is the same, but the tied version performs better when the number of parameters is kept the same, presumably due to the increased memory capacity of the state vector. Our best model has an extreme bottleneck, over a hundred of times smaller than the state size, while the word-level language models trained by \citet{Jozefowicz2016} use bottlenecks of four to eight times smaller than the state size. We conjecture that this difference is due to our usage of the "plus diagonal" parametrization. In terms of absolute perplexity, our results are worse than published ones (e.g. \citet{Graves2013a}), although they may not be directly comparable since published results generally use different training and evaluation schemes, such as preserving the network state between different sentences.

In order to address these experimental differences, we ran additional experiments using LSTM architectures, trying to replicate the alphabet and sentence segmentation used in \citet{Graves2013a}, although we could not obtain the same baseline performance even using the Adam optimizer (using SGD+momentum yields even worse results). In fact, we obtained approximately the same perplexity as our baseline GRU model with the same state size.

We applied the Low-rank plus diagonal parametrizations to our LSTM architecture maintaining the same number of parameters as the baseline. We obtained notable perplexity improvements over the baseline. Refer to appendix \ref{APPENDIXLRLSTM} for the experimental details.

We performed additional exploratory experiments on word-level language modeling and subword-level neural machine translation \citep{Bahdanau2014, Sennrich2015} with GRU-based architectures but we were not able to achieve significant accuracy improvements, which is not particularly surprising given that in these models most parameters are contained in the token embedding and output matrices, thus low-dimensional parametrizations of the recurrent matrices have little effect on the total number of parameters. We reserve experimentation on character-level neural machine translation \citep{Ling2015, Chung2016, Lee2016} to future work.

\section{Conclusions and future work}

We proposed low-dimensional parametrizations for passthrough neural networks based on low-rank  or low-rank plus diagonal decompositions of the $n \times n$ matrices that occur in the hidden layers. We experimentally compared our models with state of the art models, obtaining competitive results including a near state of the art for the randomly-permuted sequential MNIST task.

Our parametrizations are alternative to convolutional parametrizations explored by \citet{Srivastava2015, He2015, Kaiser2015}. Since our architectural innovations are orthogonal to these approaches, they can be in principle combined. Additionally, alternative parametrizations could include non-linear activation functions, similar to the network-in-network approach of \citet{Lin2013}. We leave the exploration of these extensions to future work.

{\small
\bibliography{lowrank_passthrough_nn}}

\bibliographystyle{iclr2016_conference}

\appendix

\section{Appendix: Experimental details}

\subsection{Low-rank Highway Networks}
\label{APPENDIXLRHN}

As a preliminary exploratory experiment, we applied the low-rank and low-rank plus diagonal Highway Network architecture to the classic benchmark task of handwritten digit classification on the MNIST dataset, in its permutation-invariant (i.e. non-convolutional) variant.

We used the low-rank architecture described by equations \ref{EQ:MODEL:PN:HIGHWAY} and \ref{EQ:MODEL:PN:LRHIGHWAY}, with $T=5$ hidden layers, ReLU activation function, state dimension $n = 1024$ and maximum rank (internal dimension) $d=256$. The input-to-state layer is a dense $784 \times 1024$ matrix followed by a (biased) ReLU activation and the state-to-output layer is a dense $1024 \times 10$ matrix followed by a (biased) identity activation. We did not use any convolution layer, pooling layer or data augmentation technique. We used dropout \citep{Srivastava2014} in order to achieve regularization. We further applied L2-regularization with coefficient $\lambda = \num{1e-3}$ per example on the hidden-to-output parameter matrix. We also used batch normalization \citep{Ioffe2015} after the input-to-state matrix and after each parameter matrix in the hidden layers. Initial bias vectors are all initialized at zero except for those of the transform functions in the hidden layers, which are initialized at $-1.0$. We trained to minimize the sum of the per-class L2-hinge loss plus the L2-regularization cost \citep{Tang2013}. Optimization was performed using Adam \citep{Kingma2014} with standard hyperparameters, learning rate starting at $\num{3e-3}$ halving every three epochs without validation improvements. Mini-batch size was equal to $100$. Code is available online\footnote{\url{https://github.com/Avmb/lowrank-highwaynetwork}}.

We obtained perfect training accuracy and $98.83\%$ test accuracy. While this result does not reach the state of the art for this task ($99.13\%$ test accuracy with unsupervised dimensionality reduction reported by \citet{Tang2013}), it is still relatively close. We also tested the low-rank plus diagonal Highway Network architecture of eq. \ref{EQ:MODEL:PN:LRDHIGHWAY} with the same settings as above, obtaining a test accuracy of $98.64\%$. The inclusion of diagonal parameter matrices does not seem to help in this particular task.

\subsection{Low-rank GRUs}
\label{APPENDIXLRGRU}

In our experiments (except language modeling) we optimized using RMSProp \citep{Tieleman2012} with gradient component clipping at $1$. Code is available online\footnote{\url{https://github.com/Avmb/lowrank-gru}}. Our code is based on the published uRNN code\footnote{\url{https://github.com/amarshah/complex_RNN}} (specifically, on the LSTM implementation) by the original authors for the sake of a fair comparison. In order to achieve convergence on the memory task however, we had to slightly modify the optimization procedure, specifically we changed gradient component clipping with gradient norm clipping (with NaN detection and recovery), and we added a small $\epsilon=\num{1e-8}$ term in the parameter update formula. No modifications of the original optimizer implementation were required for the other tasks.

In order to address the numerical instability issues in the memory tasks, we also consider a variant of our Low-rank plus diagonal GRU where apply weight normalization as described by \citet{Salimans2016} to all the parameter matrices except the output one and the diagonal matrices. All these matrices have trainable scale parameters, except for the projection matrices. We further apply an hard constraint on the matrices row norms by clipping them at $10$ after each update. We disable NaN detection and recovery during training.  
The rationale behind this approach, in addition to the general benefits of normalization, is that the low-rank parametrization potentially introduces stability issues because the model is invariant to multiplying a row of an $R$-matrix by a scalar $s$ and dividing the corresponding column of the $L$-matrix by $s$, which in principle allows the parameters of either matrix to grow very large in magnitude, eventually resulting in overflows or other pathological behavior. The weight row max-norm constraint can counter this problem. But the constraint alone could make the optimization problem harder by reducing and distorting the parameter space. Fortunately we could counter this by weight normalization which makes the model invariant to the row-norms of the parameter matrices.

In the language modeling experiment, for consistency with existing code, we used a variant of the GRU where the reset gate is applied after the multiplication by the recurrent proposal matrix rather than before. Specifically:
\begin{equation}
\begin{aligned}
in(u, \theta)           &= \theta_{in} \\
f_{\omega}(x(t-1), t, u, \theta)    &= \sigma(\theta^{U_{\omega}} \cdot u(t) + \theta^{(W_{\omega})} \cdot x(t-1) + \theta^{(b_{\omega})}) \\
f_{\gamma}(x(t-1), t, u, \theta)      &= \sigma(\theta^{U_{\gamma}} \cdot u(t) + \theta^{(W_{\gamma})} \cdot x(t-1) + \theta^{(b_{\gamma})}) \\
f_{\tau}(x(t-1), t, u, \theta) &=  1^{\otimes n} - f_{\gamma}(x(t-1), t, u, \theta) \\
f_{\pi}(x(t-1), t, u, \theta)    &= \text{tanh}(\theta^{U_{\pi}} \cdot u(t) + (\theta^{(W_{\pi})} \cdot x(t-1)) \odot f_{\omega}(x(t-1), t, u, \theta) + \theta^{(b_{\pi})}) 
\end{aligned}
\label{EQ:EXPERIMENTS:GRUVARIANT}
\end{equation}
The character vocabulary size if $51$, we use no character embeddings. Training is performed with Adam with learning rate $\num{1e-3}$. Bayesian recurrent dropout was adapted from the original LSTM architecture of \citet{Gal2015} to the GRU architecture as in \citet{Sennrich2016}.

Our implementation is based on the "dl4mt" tutorial\footnote{\url{https://github.com/nyu-dl/dl4mt-tutorial}} and the Nematus neural machine translation system \footnote{\url{https://github.com/rsennrich/nematus}}. The code is available online \footnote{\url{https://github.com/Avmb/dl4mt-lm/tree/master/lm}}.

\subsection{Low-rank LSTMs}
\label{APPENDIXLRLSTM}

For our LSTM experiments, we modified the implementation of LSTM language model with Bayesian recurrent dropout by \citet{Gal2015}\footnote{\url{https://github.com/yaringal/BayesianRNN}}. In order to match the setup of \citet{Graves2013a} more closely, we used a vocabulary size of $49$, no embedding layer and one LSTM layer. We found no difference on the baseline model with using peephole connections and not using them, therefore we did not use them on the Low-rank plus diagonal model. We use recurrent dropout and the Adam optimizer with learning rate $\num{2e-4}$.

The baseline LSTM model is defined by the gates:
\begin{equation}
\begin{aligned}
in(u, \theta)           &= 0^{\otimes \hat{n}} \\
f_{\omega}(x(t-1), t, u, \theta)    &= \sigma(\theta^{U_{\omega}} \cdot u(t) + \theta^{(W_{\omega})} \cdot \tilde{x}(t-1) + \theta^{(b_{\omega})}) \\
f_{\gamma}(x(t-1), t, u, \theta)      &= \sigma(\theta^{U_{\gamma}} \cdot u(t) + \theta^{(W_{\gamma})} \cdot \tilde{x}(t-1) + \theta^{(b_{\gamma})}) \\
f_{\tau}(x(t-1), t, u, \theta) &=  \sigma(\theta^{U_{\tau}} \cdot u(t) + \theta^{(W_{\tau})} \cdot \tilde{x}(t-1) + \theta^{(b_{\tau})}) \\
f_{\pi}(x(t-1), t, u, \theta)    &= \text{tanh}(\theta^{U_{\pi}} \cdot u(t) + \theta^{(W_{\pi})} \cdot \tilde{x}(t-1) + \theta^{(b_{\pi})}) \\
\end{aligned}
\label{EQ:EXPERIMENTS:LSTMGATES}
\end{equation}
with the state components evolving as:
\begin{equation}
\begin{aligned}
\hat{x}(t)   &= f_{\pi}(x(t-1), t, u, \theta) \odot f_{\tau}(x(t-1), t, u, \theta) + \hat{x}(t-1) \odot f_{\gamma}(x(t-1), t, u, \theta) \\
\tilde{x}(t)    &=  f_{\omega}(x(t-1), t, u, \theta) \odot \text{tanh}(\hat{x}(t))
\end{aligned}
\label{EQ:EXPERIMENTS:LSTMSTATE}
\end{equation}
The low-rank plus diagonal parametrization is applied on the recurrence matrices $\theta^{W_{\star}}$ as in the GRU models.

The code is available online\footnote{\url{https://github.com/Avmb/lowrank-lstm}}.
\end{document}